\pdfoutput=1

\documentclass[11pt]{article}

\usepackage{EMNLP2022}

\PassOptionsToPackage{dvipsnames}{xcolor}

\usepackage{times}
\usepackage{latexsym}
\usepackage{fdsymbol}

\usepackage[frozencache,cachedir=.]{minted}

\usepackage{listings}
\usepackage[compatibility=false]{caption}

\usepackage[T1]{fontenc}

\usepackage[utf8]{inputenc}

\usepackage{microtype}
\usepackage{booktabs}
\usepackage[normalem]{ulem}
\usepackage{paralist}
\usepackage{mathtools}
\usepackage{xspace}
\usepackage{tikz}
\usetikzlibrary{shapes,arrows}
\usetikzlibrary{positioning}
\usetikzlibrary{arrows.meta}
\usepackage{pgfplots}
\usepackage{cleveref}
\usepackage{subcaption}
\usepackage{float}
\usepackage{color}
\definecolor{deepblue}{rgb}{0,0,0.5}
\definecolor{deepred}{rgb}{0.6,0,0}
\definecolor{deepgreen}{rgb}{0,0.5,0}
\definecolor{darkgreen}{RGB}{43,163,39}
\definecolor{bluesquare}{rgb}{126,166,224}
\definecolor{LightGray}{gray}{0.9}
\definecolor{DarkGray}{gray}{0.1}

\lstdefinestyle{pythoncode}{
	language=Python,
	otherkeywords={self,join,append,split,write},   
	keywordstyle=\bfseries\color{deepblue},
	emph={__init__, digraph},          
	emphstyle=\color{deepred},    
	showstringspaces=false,
	breaklines=true,
	escapeinside=||,
	columns=fullflexible,
	basicstyle=\fontfamily{cmtt}\normalsize,
    belowskip=-\baselineskip,
    aboveskip=-0.7\baselineskip
}

\definecolor{codegreen}{rgb}{0,0.6,0}
\definecolor{codegray}{rgb}{0.5,0.5,0.5}
\definecolor{codepurple}{rgb}{0.58,0,0.82}
\definecolor{backcolour}{rgb}{0.95,0.95,0.92}

\newcommand{\eg}{\hbox{\emph{e.g.,}}\xspace}

\usepackage{enumitem}

\newcommand{\classref}[1]{\textbf{\color{blue}{#1}}}
\newcommand{\funcname}[1]{\color{brown!50!black}{#1}}
\newcommand{\varref}[1]{\color{Blue}{#1}}
\newcommand{\paren}[1]{\textbf{\color{magenta}{#1}}}

\title{Language Models of Code are Few-Shot Commonsense Learners
}

\author{Aman Madaan$^\spadesuit$, Shuyan Zhou$^\spadesuit$, Uri Alon$^\spadesuit$,\\ 
\textbf{Yiming Yang$^\spadesuit$}, \textbf{Graham Neubig$^\spadesuit$ $^\dagger$} \\
  $^\spadesuit$ Language Technologies Institute, Carnegie Mellon University, USA \\ 
  $^\dagger$ Inspired Cognition, USA \\ 
  \texttt{\{amadaan,shuyanzh,ualon,yiming,gneubig\}}@cs.cmu.edu}

\usepackage{xspace}
\usepackage{ulem}
\usepackage{soul}

\definecolor{cosmiclatte}{rgb}{1.0, 0.97, 0.91}
\definecolor{codegreen}{rgb}{0,0.6,0}
\definecolor{codegray}{rgb}{0.5,0.5,0.5}
\definecolor{codepurple}{rgb}{0.58,0,0.82}
\definecolor{backcolour}{rgb}{0.95,0.95,0.92}
\definecolor{Red}{rgb}{1,0,0}
\definecolor{Green}{rgb}{0.4,1,0.2}
\definecolor{Red}{rgb}{0.9,0,0}
\definecolor{Orange}{rgb}{1,0.5,0}
\definecolor{yellow}{rgb}{0.65,0.6,0}
\definecolor{cadmiumgreen}{rgb}{0.2, 0.7, 0.24}

\usepackage{multirow}
\newcommand{\ie}{i.e.,\xspace}
\newcommand{\code}[1]{\texttt{#1}}
\newcommand{\squishlist}{
  \begin{list}{$\bullet$}
    { \setlength{\itemsep}{0pt}      \setlength{\parsep}{3pt}
      \setlength{\topsep}{3pt}       \setlength{\partopsep}{0pt}
      \setlength{\leftmargin}{1.5em} \setlength{\labelwidth}{1em}
      \setlength{\labelsep}{0.5em} } }
\newcommand{\reallysquishlist}{
  \begin{list}{$\bullet$}
    { \setlength{\itemsep}{0pt}    \setlength{\parsep}{0pt}
      \setlength{\topsep}{0pt}     \setlength{\partopsep}{0pt}
      \setlength{\leftmargin}{0.2em} \setlength{\labelwidth}{0.2em}
      \setlength{\labelsep}{0.2em} } }

 \newcommand{\squishend}{
     \end{list} 
 }
 
\newcommand{\ours}{\textsc{CoCoGen}\xspace}
\newcommand{\genscr}{\hat{\gG}}
\newcommand{\refscr}{\gG}
\newcommand{\struct}{\gG}
\newcommand{\gene}{\hat{\mE}}
\newcommand{\refe}{\mE}

\newcommand{\kst}{\textsc{kst}\xspace}
\newcommand{\clm}{Code-LLMs\xspace}
\newcommand{\clmsingular}{Code-LLM\xspace}

\newcommand{\nlm}{NL-LLMs\xspace}

\newcommand{\dotlang}{\textsc{dot}\xspace}
\newcommand{\dotf}{\dotlang}

\newcommand{\proscr}{\textsc{proscript}\xspace}
\newcommand{\explg}{\textsc{explagraphs}\xspace}
\newcommand{\propar}{\textsc{propara}\xspace}

\newcommand{\tf}{\textsc{t5}\xspace}

\newcommand{\curie}{\textsc{curie}\xspace}
\newcommand{\codex}{\textsc{Codex}\xspace}

\newcommand{\davinci}{\textsc{davinci}\xspace}

\newcommand{\llm}{LLMs\xspace}

\newcommand{\bleu}{\textsc{bleu}\xspace}
\newcommand{\bleurt}{\textsc{bleurt}\xspace}

\newcommand{\rougel}{\textsc{rouge-l}\xspace}

\newcommand{\fscore}{$F_1$\xspace}
\newcommand{\ged}{\textsc{ged}\xspace}

\newcommand{\iso}{\textsc{iso}\xspace}
\newcommand{\avgdegree}{Avg(d)\xspace}
\newcommand{\avgnumnodes}{Avg($|\mV|$)\xspace}
\newcommand{\avgnumedges}{Avg($|\mE|$)\xspace}
\newcommand{\stca}{StCA\xspace}
\newcommand{\seca}{SeCA\xspace}
\newcommand{\gbs}{G-BS\xspace}
\newcommand{\ea}{EA\xspace}

\usepackage{amsmath,amsfonts,bm}
\usepackage{xspace}

\def\Secref#1{Section~\ref{#1}}

\def\eqref#1{equation~\ref{#1}}

\def\1{\bm{1}}

\def\mE{{\bm{E}}}

\def\mV{{\bm{V}}}

\DeclareMathAlphabet{\mathsfit}{\encodingdefault}{\sfdefault}{m}{sl}
\SetMathAlphabet{\mathsfit}{bold}{\encodingdefault}{\sfdefault}{bx}{n}

\def\gE{{\mathcal{E}}}

\def\gG{{\mathcal{G}}}

\def\gV{{\mathcal{V}}}

\newcommand{\eat}[1]{}

\newcommand{\emnlpcr}[1]{#1}

\begin{document}
\maketitle
\begin{abstract}
We address the general task of \emph{structured} commonsense reasoning: 
given a natural language input, the goal is to generate a \emph{graph} such as an event or a reasoning-graph.
To employ large language models (LMs) for this task, existing approaches ``serialize'' the output graph as a flat list of nodes and edges.
Although feasible, these serialized graphs strongly deviate from the natural language corpora that LMs were pre-trained on, hindering LMs from generating them correctly. 
In this paper, we show that when we instead frame structured commonsense reasoning tasks as \emph{code generation} tasks,
pre-trained LMs of \emph{code} are \emph{better} structured commonsense reasoners than LMs of natural language, even when the downstream task does not involve source code at all.
We demonstrate our approach across three diverse structured commonsense reasoning tasks. In all these \emph{natural language} tasks, we show that using our approach, a \textit{code} generation LM~(\codex{}) outperforms natural-LMs that are fine-tuned on the target task (\eg~\tf) and other strong LMs such as GPT-3 in the few-shot setting.
Our code and data are available at \url{https://github.com/madaan/CoCoGen} .
\end{abstract}
\section{Introduction}


The growing capabilities of large pre-trained language models (\llm) for generating text have enabled their successful application in a variety of tasks, including summarization, translation, and question-answering~\citep{wang2019superglue,raffel2019exploring,brown2020language,chowdhery2022palm}.

Nevertheless, while employing LLMs for natural language (NL) tasks is straightforward, a major remaining challenge is how to leverage LLMs for \emph{structured commonsense reasoning}, including tasks such as generating event graphs~\citep{tandon2019wiqa}, reasoning graphs~\citep{madaan2021could}, scripts~\citep{sakaguchi_proscript_2021}, and argument explanation graphs~\citep{saha_explagraphs_2021}. 
Unlike traditional commonsense reasoning tasks such as reading comprehension or question answering, \emph{structured} commonsense aims to generate \emnlpcr{structured output} given a natural language input.
This family of tasks relies on the natural language knowledge learned by the LLM,
but it also requires complex structured prediction and generation.

To leverage \llm, existing structured commonsense generation models modify the \emph{output format} of a problem.
Specifically, the structure to be generated~(\eg a graph or a table) is converted, or ``serialized'', into text.
Such conversions include ``flattening'' the graph into a list of node pairs (\Cref{fig:mainfig:list}),
or into a specification language such as \dotf~\citep[\Cref{fig:mainfig:dot}; ][]{gansner2006drawing}.

\begin{figure*}[!h]
\centering
  \begin{subfigure}[b]{.48\linewidth}
  \raisebox{10mm}{
  \begin{tikzpicture}[node distance=1.5cm]
\newcommand{\nodespacing}{3pt}
\newcommand{\roundness}{2mm}
\newcommand{\intromarkersize}{1.5mm}
    
    \node [draw, rounded corners=\roundness,inner sep=\nodespacing] (node0) {Take the pies out to cool};
    \node [draw, right of=node0, xshift=2.5cm,rounded corners=\roundness,inner sep=\nodespacing]  (node1) {Open cabinet drawer};
    \node [draw, below of=node0, xshift=2cm,rounded corners=\roundness,inner sep=\nodespacing] (node2) {Take out several plates};
    \node [draw, below of=node2, xshift=-2cm,rounded corners=\roundness*2,inner sep=\nodespacing, text width=3cm,align=center] (node3) {Begin putting pies on plate};
    \node [draw, right of=node3, xshift=2.5cm,rounded corners=\roundness*2,inner sep=\nodespacing, text width=3cm,align=center] (node4) {Fill pies onto plates evenly};
    \node [draw, below of=node3, xshift=2cm,rounded corners=\roundness,inner sep=\nodespacing] (node5) {Serve the potpies on a plate};
    \draw [-{Latex[length=\intromarkersize, line width=1pt]}] (node0) -- (node2);
    \draw [-{Latex[length=\intromarkersize, line width=1pt]}] (node1) -- (node2);
    \draw [-{Latex[length=\intromarkersize, line width=1pt]}] (node2) -- (node3);
    \draw [-{Latex[length=\intromarkersize, line width=1pt]}] (node2) -- (node4);
    \draw [-{Latex[length=\intromarkersize, line width=1pt]}] (node3) -- (node5);
    \draw [-{Latex[length=\intromarkersize, line width=1pt]}] (node4) -- (node5);
\end{tikzpicture}}
  \caption{The script $\struct$}
  \label{fig:mainfig:graph}
  \end{subfigure}
  \hfill
   \begin{subfigure}[b]{.49\linewidth}
   \centering
 \begin{minted}[fontsize=\footnotesize,framesep=1pt,frame=single,autogobble,breaklines,breaksymbolleft=\;,escapeinside=||]{python}
class Tree:

  |\varref{goal}| = "serve the potpies on a plate"

  def |\funcname{\_\_init\_\_}||\paren{(}||\varref{self}||\paren{)}|:
    # nodes
    |\varref{take\_pies\_out\_to\_cool}| = |\classref{Node}||\paren{()}|
    |\varref{open\_cabinet\_drawer}| = |\classref{Node}||\paren{()}|
    |\varref{take\_out\_several\_plates}| = |\classref{Node}||\paren{()}|
    ...
    # edges
    |\varref{take\_pies\_out\_to\_cool}|.children = |\paren{[}||\varref{take\_out\_several\_plates}||\paren{]}|
    |\varref{open\_cabinet\_drawer}|.children = |\paren{[}||\varref{take\_out\_several\_plates}||\paren{]}|
    ...
   \end{minted}
     \caption{$\struct$ converted to Python code $\struct_c$ using our approach}
  \label{fig:mainfig:code}
    \end{subfigure}
  \vskip\baselineskip
    \begin{subfigure}[b]{.49\linewidth}
    \centering
 \begin{minted}[fontsize=\footnotesize,tabsize=2,breaklines,breaksymbolleft=\;,framesep=1pt,frame=single]{text}
digraph G {
  begin -> take_pies_out_to_cool;
  begin -> open_cabinet_drawer;
  take_pies_out_to_cool -> take_out_several_plates;
  open_cabinet_drawer -> take_out_several_plates;
  take_out_several_plates -> begin_putting_pies_on_plates;
  begin_putting_pies_on_plates -> serve_potpies_on_plate;
  fill_pies_onto_plates_evenly -> serve_potpies_on_plate;
  serve_potpies_on_plate -> end;
}
   \end{minted}
     \caption{Straightforward encodings of the graph using the ``DOT''}
  \label{fig:mainfig:dot}
   \end{subfigure}
  \hfill
  \begin{subfigure}[b]{.49\linewidth}
   \centering
 \begin{minted}[fontsize=\footnotesize,tabsize=2,breaklines,breaksymbolleft=\;,framesep=1pt,frame=single]{text}
[
  (take_pies_out_to_cool, take_out_several_plates), 
  (open_cabinet_drawer, take_out_several_plates),
  (take_out_several_plates, begin_putting_pies_on_plates),
  (take_out_several_plates, fill_pies_onto_plates_evenly),
  (begin_putting_pies_on_plates, serve_potpies_on_plate),
  (fill_pies_onto_plates_evenly, serve_potpies_on_plate),
  (serve_potpies_on_plate, end)
]
   \end{minted}
        \caption{\raggedright Text format, or as a list of edges (node pairs)}
  \label{fig:mainfig:list}
    \end{subfigure}
 \caption{An illustration of \ours for the task of script generation. An input graph (\ref{fig:mainfig:graph}) is typically represented using the DOT format (\ref{fig:mainfig:dot}) or as a list of edges (\ref{fig:mainfig:list}), which allows modeling the graph using standard language models. These popular choices are sufficient in principle; however, these formats are loosely structured, verbose, and not common in text corpora, precluding language models from effectively generating them. In contrast, \ours converts structures into Python code (\ref{fig:mainfig:code}), allowing to model them using large-scale language models of \emph{code}.
 }
\label{fig:mainfigure}
 \end{figure*}

While converting the structured output into text 
has shown promising results \cite{rajagopal2021curie,madaan2020neural}, \llm struggle to generate these  ``unnatural'' outputs: LMs are primarily pre-trained on free-form text, and these serialized structured outputs strongly diverge from the majority of the pre-training data.
Further, for natural language, semantically relevant words are typically found within a small span, whereas neighboring nodes in a graph might be pushed farther apart when representing a graph as a flat string.

Thus, a language model which was trained on natural language text is likely to fail to capture the topology of the graph. 
Consequently, using \llm for graph generation typically requires a large amount of task-specific training data, and their generated outputs show structural errors and semantic inconsistencies, which need to be further fixed either manually or by using a secondary downstream model~\citep{madaan2021think}.

Despite these struggles, the recent success of large-language models of \emph{code}~\citep[Code-LLMs; ][]{chen2021evaluating,xu2022systematic} for tasks such as code generation from natural language \cite{austin2021program,Nijkamp2022ACP}, code completion \cite{fried2022incoder}, and code translation \cite{wang2021codet5}, show that Code-LLMs are able to perform complex reasoning on structured data such as programs. 
Thus, instead of forcing \llm of natural language (\nlm) to be fine-tuned on structured commonsense data, an easier way to close the discrepancy between the pre-training data (free-form \emph{text}) and the task-specific data (commonsense reasoning \emph{graphs}) is to adapt LLMs that were pre-trained on \emph{code} to structured commonsense reasoning in \emph{natural language}.


Thus, our main insight is that \emph{large language models of code are good structured commonsense reasoners}. Further, we show that \clm can be even better structured reasoners than \nlm, when converting the desired output graph into a format similar to that observed in the code pre-training data.
We call our method \ours{}: models of \textbf{Co}de for \textbf{Co}mmonsense \textbf{Gen}eration, and it is demonstrated in \Cref{fig:mainfigure}.

Our contributions are as follows:
\begin{compactenum}
    \item 
We highlight the insight that \clm are better structured commonsense reasoners than \nlm, when representing the desired graph prediction as code.
    \item We propose \ours: a method for leveraging LLMs of \textbf{co}de for structured \textbf{co}mmonsense \textbf{gen}eration.
    \item We perform an extensive evaluation across three  structured commonsense generation tasks and demonstrate that \ours{} vastly outperforms NL-\llm, either fine-tuned or few-shot tested, while controlling for the number of downstream task examples. 
    \item We perform a thorough ablation study, which shows the role of data formatting, model size, and the number of few-shot examples.
\end{compactenum}

\section{\ours: Representing Commonsense structures with code}
\label{sec:method}

We focus on tasks of structured commonsense generation.
Each training example for such tasks is in the form $(\mathcal{T}, \struct)$, where $\mathcal{T}$ is a text input, and $\struct$ is the structure to be generated (typically a graph).
The key idea of \ours is transforming an output graph $\struct$ into a semantically equivalent program $\struct_{c}$ written in a general-purpose programming language. 
In this work, we chose Python due to its popularity in the training data of modern \clm~\citep{xu2022systematic}, but our approach is agnostic to the programming language. 
The code-transformed graphs are similar in their format to the pre-training data of \clm, and thus serve as easier to generalize training or few-shot examples than the original raw graph. 
\ours uses \clm to generate $\struct_{c}$ given $\mathcal{T}$, which we eventually convert back into the graph $\struct$.

We use the task of script generation~(\proscr, Figure~\ref{fig:mainfigure}) as a running example to motivate our method: 
script generation aims to create a script~($\struct$) to achieve a given high-level goal ($\mathcal{T}$).

\subsection{Converting $(\mathcal{T}, \struct)$ into Python code} 
We convert a $(\mathcal{T}, \struct)$ pair into a Python class or function.
The general procedure involves adding the input text $\mathcal{T}$ in the beginning of the code as a class attribute or descriptive comment, and encoding the structure $\struct$ using standard constructs for representing structure in code~(\eg~hashmaps, object attributes) or function calls.
The goal here is to compose Python code that represents a $(\mathcal{T}, \struct)$ pair, but retains the syntax and code conventions of typical Python code.

For example, for the script generation task, we convert the $(\mathcal{T}, \struct)$ pair into a \code{Tree} class~(\Cref{fig:mainfig:code}).
The goal $\mathcal{T}$ is added as class attribute (\code{goal}), and the script $\struct$ is added by listing the nodes and edges separately.
We first instantiate the list of nodes as objects of class \code{Node}.
Then, the edges are added as an attribute \code{children} for each node~(\Cref{fig:mainfig:code}).
For example, we instantiate the node ``\emph{Take out several plates}'' as  \code{take\_out\_several\_plates = Node()}, and add it as a child of the node \code{take\_pies\_out\_to\_cool}.

While there are multiple ways of representing a training example as a Python class, we found empirically that this relatively simple format is the most effective, especially with larger models.
We analyze the choice of format and its connection with the model size in \Secref{sec:analysis}.

\subsection{Few-shot prompting for generating $\struct$}
We focus on large-language models of the scale of \codex~\citep{chen_evaluating_2021}. Due to their prohibitively expensive cost to fine-tune, these large models are typically used in a \textit{few-shot prompting} mode.
Few-shot prompting uses $k$ input-output examples $\{(x_i, y_i)\}_{i=1}^{k}$ to create an in-context prompt: $p = x_1\oplus y_1 \; \cdot \; x_2\oplus y_2\; \cdot\; \ldots \cdot\; x_k\oplus y_k$, 
where $\oplus$ is a symbol that separates an input from its output, and $\cdot$ separates different examples.

A new (test) input $x$ is appended to the prompt $p$ (that is: $p\; \cdot\; x$), and $p\; \cdot\; x\; \oplus$ is fed to the model for completion.
As found by~\citet{brown2020language}, large language models show impressive few-shot capabilities in generating a completion $\hat{y}$ given the input $p\; \cdot\; x \; \oplus$. The main question is how to construct the prompt?

In all experiments in this work, the prompt $p$ consists of $k$ Python classes, each representing a $(\mathcal{T}, \struct_c)$ pair.
For example, for script generation, each Python class represents a goal $\mathcal{T}$ and a script $\struct_c$ from the training set.
Given a new goal $\mathcal{T}$ for inference, a partial Python class (\ie only specifying the goal) is created and appended to the prompt.
\Cref{listing:inferenceinput} shows such a partial class. Here, the code generation model is expected to complete the class by generating the definition for \code{Node} objects and their dependencies for the goal  \textit{make hot green tea}.

\begin{figure}[!h]
\centering
 \begin{minted}[fontsize=\normalsize,framesep=1pt,frame=single,autogobble,breaklines,breaksymbolleft=\;,escapeinside=||]{python}
class Tree:
  |\varref{goal}| = "make hot green tea."

  def |\funcname{\_\_init\_\_}||\paren{(}||\varref{self}||\paren{)}|:
    # generate
\end{minted}
\caption{\ours uses a prompt consisting of $k$~(5-10) Python classes. During inference, the test input is converted to a partial class, as shown above, appended to the prompt, and completed by a code generation model such as \codex.}
\label{listing:inferenceinput}
\end{figure}

In our experiments, we used \codex~\citep{chen_evaluating_2021} and found that it nearly always generates syntactically valid Python.
Thus, the generated code can be easily converted back into a graph and evaluated using the dataset's standard, original, metrics. Appendix~\ref{sec:appendixprompts} lists sample prompts for each of the tasks we experimented with.

 
 
 


\section{Evaluation}
\label{sec:evaluation}

We experiment with three diverse structured commonsense generation tasks: 
\begin{inparaenum}[(i)]
\item script generation~(\proscr, \Secref{sec:scriptgen}),
\item entity state tracking~(\propar, \Secref{sec:propara}), and
\item explanation graph generation~(\explg, \Secref{sec:explagraphs})
\end{inparaenum}
Dataset details are included in~\Cref{sec:appendixdataset}. Despite sharing the general goal of structured commonsense generation, the three tasks are quite diverse in terms of the generated output and the kind of required reasoning.

\subsection{Experimental setup}

\paragraph{Model} 
As our main \clmsingular for~\ours, we experiment with the latest version of \codex \code{code-davinci-002} from OpenAI\footnote{As of June 2022} in few-shot prompting mode.

\paragraph{Baselines} 
We experimented with the following types of baselines:
\begin{enumerate}
    \item \textbf{Text few-shot:} Our hypothesis is that code-generation models can be repurposed to generate structured output better. Thus,  natural baselines for our approach are \nlm~-- language models trained on natural language corpus. 
    We experiment with the latest versions of \curie~(\code{text-curie-001}) and \davinci~(\code{text-davinci-002}), the two GPT-3 based models by OpenAI~\citep{brown2020language}.
    For both these models, the prompt consists of $(\mathcal{T}, \struct)$ examples, where $\struct$ is simply flattened into a string~(as in \Cref{fig:mainfig:dot}).
    \davinci is estimated to be much larger in size than \curie, as our experiments also reveal~(Appendix~\ref{sec:modeldescription}).
    \davinci, popularly known as GPT-3, is the strongest text-generation model available through OpenAI APIs.\footnote{\url{https://beta.openai.com/docs/models/gpt-3}}

    \item \textbf{Fine-tuning:} we fine-tune a \tf-large model for \explg, and use the results from \citet{sakaguchi_proscript_2021} on \tf-xxl for \proscr tasks. In contrast to the few-shot setup where the model only has access to a few examples, fine-tuned models observe the \emph{entire} training data of the downstream task.
\end{enumerate}

\paragraph{Choice of prompt} We created the prompt $p$  by randomly sampling $k$ examples from the training set. As all models have a bounded input size (\eg 4096 tokens for  \codex \code{code-davinci-002} and 4000 for  GPT-3 \code{text-davinci-002}),
the exact value of $k$ is task dependent: more examples can fit in a prompt in tasks where $(\mathcal{T}, \struct)$ is short.
In our experiments, $k$ varies between $5$ and $30$, and the GPT-3 baseline is always fairly given the same prompts as \codex.
To control for the variance caused by the specific examples selected into $p$, we repeat each experiment with at least 3 different prompts, and report the average.
We report the mean and standard deviations in \Cref{sec:variationprompts}.

\textbf{\ours:} We use \ours to refer to setups where a \codex is used with a Python prompt.
In ~\Secref{sec:analysis}, we also experiment with dynamically creating a prompt for each input example, using a \nlm with code prompts, and using \clm with textual prompts.

\begin{table*}[]
    
    \centering
    \begin{tabular}{lrrrrrrrr}
        \toprule
        & \bleu & \rougel & \bleurt & \iso & \ged  & \avgdegree & \avgnumnodes & \avgnumedges \\ 
        \midrule
        \multicolumn{2}{l}{$\refscr$ (reference graph)} & - & - & 1.00  & 0.00      & 1.84       & 7.41         & 6.80          \\
        \midrule
        \tf (fine-tuned)    & 23.80  & 35.50      & -0.31 & 0.51& 1.89 &  \textbf{1.79} & 7.46 & \textbf{6.70}  \\
        \midrule
        \curie~(15) & 11.40  & 27.00       & -0.41 & 0.15 & 3.92  & 1.47 & 8.09 & 6.16 \\
    \davinci~(15)  &   23.11& 	36.51 &	-0.27 & \textbf{0.64} &  	\textbf{1.44} &  	1.74&  	7.58&  	6.59              \\
    
    \ours~(15)    & \textbf{25.24}	& \textbf{38.28} &	\textbf{-0.26 } &   0.53 &	2.10	& \textbf{1.79} &	\textbf{7.44} &	\textbf{6.70}              \\ 
    \bottomrule
\end{tabular}
\caption{Semantic and structural metrics for the script generation task on $\proscr$. \tf is fine-tuned on the entire dataset, while the few-shot models~(\curie, \davinci, \codex) use 15 examples in the prompt.
}
\label{tab:proscriptscriptgenresults}
\end{table*}

\begin{table}[!ht]
\centering
\small
\begin{tabular}{llccc}
\toprule
    & Method               & $prec$ & $rec$ & \fscore \\ \midrule
\multirow{3}{*}{fine-tuned} & \tf (100)         & 52.26 & 52.91 & 51.89   \\
& \tf (1k)          & 60.55 & 61.24 & 60.15   \\
 & \tf (4k)          & \textbf{75.71} & \textbf{75.93} & \textbf{75.72}   \\ 
\midrule
\multirow{3}{*}{few-shot} & \curie~(15)            & 10.19 & 11.61 & 10.62   \\
 & \davinci~(15)          & 50.62 & 49.30 & 48.92   \\
 & \ours~(15)    & \textbf{57.34}& \textbf{55.44} &  \textbf{56.24}   \\ \bottomrule
\end{tabular}
\caption{Precision, recall, and \fscore for $\proscr$ edge-prediction task. \ours with 15 samples outperforms strong few-shot models, and \tf trained on 100 samples.}
\label{tab:proscriptedgepredresults}
\end{table}

\subsection{Script generation: \proscr}
\label{sec:scriptgen}
Given a high-level goal (\eg \textit{bake a cake}), 
the goal of script generation is to generate a graph where each node is an action, and edges capture dependency between the actions (\Cref{fig:mainfig:graph}). 
We use the \proscr~\citep{sakaguchi_proscript_2021} dataset, where the scripts are directed acyclic graphs, which were collected from a diverse range of sources including ROCStories~\citep{mostafazadeh2016corpus}, Descript~\citep{wanzare2016crowdsourced}, and Virtual home~\citep{puig2018virtualhome}.

Let $\refscr(\gV, \gE)$ be a script for a high-level goal $\mathcal{T}$ with node and edge sets $\gV$ and $\gE$, respectively.
Following \citet{sakaguchi_proscript_2021}, we experiment with two sub-tasks: 
\begin{inparaenum}[(i)]
    \item \textbf{script generation:} generating the entire script $\refscr(\gV, \gE)$ given a goal $\mathcal{T}$, and 
    \item \textbf{edge prediction:} predicting the edge set $\gE$ \textbf{given} the nodes $\gV$ and the goal $\mathcal{T}$.
\end{inparaenum}

\Cref{fig:mainfigure} shows an input-output example from \proscr, and our conversion of the graph into Python code:
we convert each node $v \in \gV$ into an instance of a \code{Node} class; we create the edges by adding \code{children} attribute for each of the nodes. 
Additional examples are present in Figure~\ref{fig:proscript_example_complete}

To represent a sample for edge prediction, we list the nodes in a random order (specified after the comment~\emph{\code{\# nodes}} in \Cref{fig:mainfig:code}).
The model then completes the class by generating the code below the comment~\emph{\code{\# edges}}.

\paragraph{Script Generation metrics}
We denote the script that was generated by the model as  $\genscr$, and evaluate $\genscr$ vs. $\refscr$ for both semantic and structural similarity.
To evaluate semantic similarity, 
we use \bleu, \rougel, and the learned metric \bleurt to determine the content overlap. Following \citet{sakaguchi_proscript_2021}, we use the following metrics for structural evaluation of generated scripts:
\begin{itemize}[noitemsep,topsep=0pt,parsep=0pt,partopsep=0pt]
    \item Graph edit distance (\ged): the number of required edits (node/edge removal/additions)  to transform $\genscr$ to $\refscr$ \cite{abu2015exact}; 
    \item Graph isomorphism \cite[\iso; ][]{cordella2001improved}: determines whether  $\genscr$ and $\refscr$ are isomorphic based on their structure,  disregarding the textual content of nodes;
    \item Graph size: average number of nodes and edges,  $(|\refscr(V)| ,|\refscr(E)|, |\genscr(V)|, |\genscr(V))$ and the average degree~($\text{d}(\refscr(V))$), where the high-level goal is for $\genscr$ to have as close  measures to $\refscr$ as possible.
\end{itemize}

\paragraph{Edge Prediction metrics}
For the edge prediction task,  the set of nodes is given, and the goal is to predict the edges between them.
Following~\citet{sakaguchi_proscript_2021}, we measure precision, recall, and \fscore comparing the true and predicted edges.
Specifically, $p = \frac{|\refe \cap \gene|}{|\gene|}$, $r = \frac{|\refe \cup \gene|}{|\refe|}$, and $F_1 = \frac{2pr}{p + r}$.

\begin{figure*}[!h]
  \begin{minipage}{0.36\textwidth}
   \centering
         \centering
         \begin{tabular}{cccc} 
         \toprule
          Action & \multicolumn{3}{c}{Entity} \\
          \midrule
           &  water & light & CO2 \\
          \midrule
          Initial states &  soil & sun & - \\
          \midrule
          Roots absorb  \\
          water from soil &  roots & sun & ? \\
          \midrule
          The water flows \\
          to the leaf &  leaf & sun & ?  \\
          
          \bottomrule 
         \end{tabular}
  \end{minipage}
  \hfill
  \begin{minipage}{0.54\textwidth}
   \centering
 \begin{minted}[fontsize=\footnotesize,framesep=1pt,frame=single,autogobble,escapeinside=||]{python}
 def |\funcname{main}||\paren{()}|:
    # init
    # roots absorb water from soil
    # the water flows to the leaf
    # state_0 tracks the location/state water
    # state_1 tracks the location/state light
    # state_2 tracks the location/state CO2
    def |\funcname{init}||\paren{()}|:
      |\varref{state\_0}| = "soil"
      |\varref{state\_1}| = "sun"
      |\varref{state\_2}| = None
    def |\funcname{roots\_absorb\_water\_from\_soil}||\paren{()}|:
      |\varref{state\_0}| = "roots"
      |\varref{state\_1}| = "sun"
      |\varref{state\_2}| = "UNK"
    def |\funcname{water\_flows\_to\_leaf}||\paren{()}|:
      |\varref{state\_0}| = "leaf"
      |\varref{state\_1}| = "sun"
      |\varref{state\_2}| = "UNK"
   \end{minted}
 
  \end{minipage}
  \captionof{figure}{A \propar example (left) and its corresponding Python code (right). We use a string to represent a concrete location~(\eg~\code{soil}), \code{UNK} to represent an unknown location, and \code{None} to represent non-existence.}
   \label{lst:propara_example}
 \end{figure*}

\paragraph{Results}
\Cref{tab:proscriptscriptgenresults} shows the results for script generation. 
The main results are that \ours (based on \codex), with just 15 prompt examples, outperforms the fine-tuned model \tf which has been fine-tuned on \emph{all} 3500 samples. Further, \ours outperforms the few-shot NL-LM \curie{} across all semantic metrics and structural metrics. \ours{} outperforms \davinci across all semantic metrics, while \davinci performs slightly better in two structural metrics.

\Cref{tab:proscriptedgepredresults} shows the results for edge prediction: \ours{} significantly outperforms the NL-LLMs \curie{} and \davinci{}.
When comparing with \tf{}, which was fine-tuned,
\ours with only 15 examples outperforms the fine-tuned \tf which was fine-tuned on 100 examples.
The impressive performance in the edge-generation task allows us to highlight the better ability of \ours{} in capturing structure,  while factoring out all models' ability to generate the NL content.


 \subsection{Entity state tracking: \propar}
\label{sec:propara}
The text inputs $\mathcal{T}$ of entity state tracking are a sequence of actions in natural language about a particular topic~(\eg~photosynthesis) and a collection of entities~(\eg~water). The goal is to predict the state of each entity after the executions of an action.  
We use the \propar dataset~\cite{mishra_tracking_2018} as the test-bed for this task. 
 
We construct the Python code $\mathcal{G}_c$ as follows, and an example is shown in~\autoref{lst:propara_example}. 
First, we define the \code{main} function and list all $n$ actions as comments inside the \code{main} function. 
Second, we create $k$ variables named as \code{state\_k} where $k$ is the number of participants of the topic. 
The semantics of each variable is described in the comments as well. 
Finally, to represent the state change after each step, we define $n$ functions where each function corresponds to an action. 
We additionally define an \code{init} function to represent the initialization of entity states.
Inside each function, the value of each variable tells the state of the corresponding entity after the execution of that action. Given a new test example where only the actions and the entities are give, we construct the input string until the \code{init} function, and we append it to the few-shot prompts for predictions.

 \paragraph{Metrics}
We follow~\citet{mishra_tracking_2018} and measure precision, recall and \fscore{} score of the predicted entity states.
We randomly sampled three examples from the training set as the few-shot prompt.
 
\paragraph{Results}

\begin{table}[]
    \centering
    \begin{tabular}{lccc}
    \toprule
    Model & $prec$ & $rec$ & $F_{1}$ \\
    \midrule
    \curie & \textbf{95.1} & 22.3  & 36.1 \\
    \davinci & 75.5 & 47.1 & 58.0\\
    \ours & 80.0 & \textbf{53.6} & \textbf{63.0}\\
    \bottomrule
    \end{tabular}
    \caption{3-shots results on \propar. All numbers are averaged among five runs with different randomly sampled prompts. 
\ours significantly outperforms \curie and \davinci. 
}
    \label{tab:propararesults_curie}
\end{table}


As shown in \Cref{tab:propararesults_curie}, \ours{} achieves a significantly better \fscore{} score than \davinci. 
Across the five prompts, \ours{} achieves 5.0 higher \fscore than \davinci on average.
In addition, \ours{} yields stronger performance than \curie, achieving \fscore of 63.0, which is 74\% higher than \curie (36.1).\footnote{\curie often failed to produce output with the desired format, and thus its high precision and low recall.}

In \propar, \ours will be ranked $6^{th}$ on the leaderboard.\footnote{As of 10/11/2022, \url{https://leaderboard.allenai.org/propara/submissions/public}}
However, all the methods above \ours require fine-tuning on the entire training corpus. In contrast, \ours uses only \textit{3 examples} in the prompt and has a gap of less than 10 \fscore points vs. the current state-of-the-art~\citep{ma2022coalescing}.
In the few-shot settings, \ours is state-of-the-art in \propar.
\begin{figure*}[!h]
 \begin{minipage}{0.40\textwidth}
  \begin{tikzpicture}[node distance=2.2cm]
    \centering
        \node [ellipse,draw=brown!80!white,inner sep=2pt,fill=Apricot!20!white,line width=1pt,text width=1.5cm,align=center] (node0) {Factory Farming};
        \node [ellipse, draw=OliveGreen!80!white,right of=node0, xshift=1.8cm,inner sep=2pt,fill=OliveGreen!10!white,line width=1pt,align=center]  (node1) {Millions};
        \node [ellipse, draw=Maroon, dashed, below of=node1, xshift=0cm,inner sep=2pt,fill=Maroon!10!white,line width=1pt,align=center] (node2) {Food};
        \node [ellipse, draw=Maroon, dashed, below of=node0, xshift=0cm, yshift=0cm, inner sep=2pt,fill=Maroon!10!white,line width=1pt,align=center] (node3) {Necessary};
        \node [ellipse, draw=NavyBlue, below of=node3,yshift=0cm, inner sep=2pt,fill=NavyBlue!10!white,line width=1pt,align=center] (node4) {Banned};
        \draw [-{Latex[length=3mm, line width=1pt]},out=-30,in=150] (node0.east) to node[pos=0.2,fill=white,xshift=2mm] {causes} (node2.north) ;
        \draw [-{Latex[length=3mm, line width=1pt]}] (node0) -- (node3) node[pos=0.4,fill=white] {has context};
        \draw [-{Latex[length=3mm, line width=1pt]}] (node1.south) to node[midway,fill=white] {desires} (node2.north);
        \draw [-{Latex[length=3mm, line width=1pt]},out=180,in=0] (node2.west) to node[midway,below] {has context} (node3.east) ;
        \draw [-{Latex[length=3mm, line width=1pt]}] (node3) -- (node4) node[midway,fill=white] {not desires};
    \end{tikzpicture}
\end{minipage}\hfill
 \begin{minipage}{0.60\textwidth}
  \centering
\begin{minted}[fontsize=\footnotesize,framesep=1pt,frame=single,autogobble,tabsize=2,breaklines,breaksymbolleft=\;,escapeinside=||]{python}
class ExplanationDAG:

  def |\funcname{\_\_init\_\_}||\paren{(}||\varref{self}||\paren{)}|:
    |\varref{belief}| = "factory farming should not be banned."
    |\varref{argument}| = "Factory farming feeds millions."
    |\varref{stance}| = "support"

    # Edges
    |\varref{begin}| = |\paren{[}|"factory farming", "millions"|\paren{]}|
    |\funcname{add\_edge}||\paren{(}|"factory farming", "causes", "food"|\paren{)}|
    |\funcname{add\_edge}||\paren{(}|"factory farming", "has context", "necessary"|\paren{)}|
    |\funcname{add\_edge}||\paren{(}|"food", "has context", "necessary"|\paren{)}|
    |\funcname{add\_edge}||\paren{(}|"necessary", "not desires", "banned"|\paren{)}|
    |\funcname{add\_edge}||\paren{(}|"millions", "desires", "food"|\paren{)}|
  \end{minted}

 \end{minipage}
 \captionof{figure}{An explanation graph (left) and the corresponding Python code (right)} 
  \label{fig:explagraph_example}
\end{figure*}
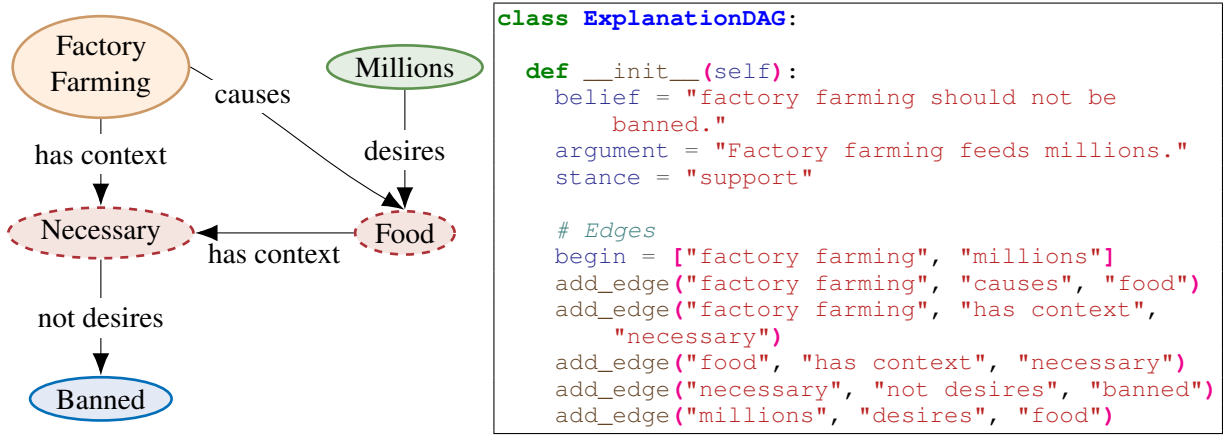
\begin{table*}[]
\centering
\begin{tabular}{llrrrrr}
\toprule
            & & \stca~($\uparrow$) & \seca~($\uparrow$) & \gbs~($\uparrow$)  & \ged~($\downarrow$)  & \ea~($\uparrow$)   \\ \midrule
\multirow{2}{*}{fine-tuned} & \tf (150)   & 12.56 & 6.03  & 9.54  & 91.06 & 7.77   \\
& \tf (1500)  & 38.19 & 21.86 & 29.37 & 73.09 & 23.41  \\
& \tf (2500)  & 43.22 & \textbf{29.65} & 33.71 & 69.14 & \textbf{26.38}  \\ 
\midrule
\multirow{3}{*}{few-shot} &\curie (30)     & 5.03  & 1.26  & 3.95  & 96.74 & 2.60   \\
& \davinci (30)    &  23.62 &	10.80	& 18.46	&83.83	&11.84     \\
& \ours (30) & \textbf{45.20}  & \textbf{23.74} & \textbf{34.68} & \textbf{68.76} & \textbf{23.58}  \\ \bottomrule
\end{tabular}
\caption{Results for \explg~(eval split). 
\ours with only 30 examples outperforms the \tf model which was fine-tuned on 1500 examples, across all metrics.}
\label{tab:explagraphresults}
\end{table*}

\subsection{Argument graph generation: \explg}
\label{sec:explagraphs}
Given a belief (\eg \textit{factory farming should not be banned}) and an argument (\eg \textit{factory farming feeds millions}), the goal of this task is to generate a graph that uses the argument to either \textit{support} or \textit{counter} the belief~\citep{saha_explagraphs_2021}.
The text input to the task is thus a tuple of (\emph{belief}, \emph{argument}, \emph{``supports''/``counters''}), and the structured output is an explanation graph~(Figure~\ref{fig:explagraph_example}).

We use the \explg dataset for this task~\citep{saha_explagraphs_2021}.
Since we focus on generating the argument graph, we take the stance as given and use the stance that was predicted by a stance prediction model released by~\citeauthor{saha_explagraphs_2021}.

To convert an \explg to Python, the belief, argument, and stance are instantiated as string variables.
Next, we define the graph structure by specifying the edges.
Unlike \proscr, the edges in \explg are typed.
Thus, each edge is added as an \code{add\_edge(source, edge\_type, destination)} function call.
We also list the starting nodes in a list instantiated with a \code{begin} variable~(Figure~\ref{fig:explagraph_example}).
Given a test example, we construct the input until the line of \code{\#\; Edges} and let a model complete the remaining.

\paragraph{Metrics}
We use the metrics defined by~\citet{saha_explagraphs_2021} 
(see Section 6 of ~\citet{saha_explagraphs_2021} for a detailed description of the mechanisms used to calculate these metrics):
\squishlist
    \item Structural accuracy~(\stca): fraction of graphs that are connected DAGs with two concepts each from belief and the argument.
    \item Semantic correctness~(\seca): a learned metric that evaluates if the correct stance is inferred from a (belief, graph) pair.
    \item G-BERTScore~(\gbs): measures BERTscore-~\citep{zhang2019bertscore} based overlap between generated and reference edges .
    \item GED~(\ged): avg. edits required to transform the generated graph to the reference graph.
    \item Edge importance accuracy~(\ea): measures the importance of each edge in predicting the target stance. A high \ea implies that each edge in the generated output contains unique semantic information, and removing any edge will hurt.
\squishend

\paragraph{Results}
\Cref{tab:explagraphresults} shows that 
\ours with only 30 examples outperforms the \tf model that was fine-tuned using 1500 examples, 
across all metrics.
Further, \ours outperforms the \nlm  \davinci and \curie with a text-prompt across all metrics by about 50\%-100\%.

\section{Analysis}
\label{sec:analysis}

In this section, we analyze the effect of three important components of \ours:
\begin{inparaenum}[(i)]
\item the contributions of \clm and structured prompt $\struct_c$;
\item the selection of examples in the in-context prompt; and
\item the design of the Python class.

\end{inparaenum}

\begin{table*}
\centering
\small
\begin{tabular}{lrrrrrrrr}
\toprule
& \multicolumn{5}{c}{\explg} & \multicolumn{3}{c}{\proscr (edge-prediction)} \\
            & \stca~($\uparrow$) & \seca~($\uparrow$) & \gbs~($\uparrow$)  & \ged~($\downarrow$)  & \ea~($\uparrow$) & $p$ & $r$ & \fscore   \\ \midrule
\davinci + text & \textbf{33.16} & 	7.14& 	25.91& 	77.45& 	15.9 & 43.06 & 41.52 & 43.06 \\
\davinci + code & 33.00 &	\textbf{15.37} &	\textbf{26.15} &	\textbf{76.91} &	\textbf{16.68} & \textbf{50.62} & \textbf{48.27} & \textbf{49.3} \\ \midrule 
 \codex  + text & 38.02 &	18.23 &	29.46 &	73.68 &	19.54 & 45.31 & 43.95 & 44.47\\
\ours~(\codex + code) & \textbf{45.20}& 	\textbf{23.74}& 	\textbf{34.68}& 	\textbf{68.76}& 	\textbf{23.58}  & \textbf{57.34} & \textbf{55.44} & \textbf{56.52} \\ \bottomrule
\end{tabular}
\caption{Teasing apart the contributions of a code generation model and a structured prompt. The experiments show that both are helpful. \davinci, a text generation model, shows marginal improvements with a code prompt (top two rows). Similarly, \codex, a code generation model, significantly benefits from a code prompt. Overall, \codex with code prompt performs better than the alternatives, across all metrics.}
\label{tab:structurevstext}
\end{table*}

\paragraph{Structured Prompts vs. \clm}
Which component is more important, using a \clm or the structured formatting of the input as code?
To answer this, we experimented with a text prompt with a Code-LLM \codex, and a code prompt with an NL-LLM, \davinci.
\Cref{tab:structurevstext} shows that both contributions are indeed important: performance improves for the NL-LLM \davinci both when we use a code prompt, \emph{and} when we use a Code-LLM. However when using both a Code-LLM and a code prompt -- the improvement is greater than the sum of each of these solely.

\paragraph{Dynamic prompt selection}
The prompts for all experiments in \Cref{sec:evaluation} were created by \emph{random} sampling of examples from the training set.
Specifically, a set of $k$ $(\mathcal{T}, \struct)$ pairs are sampled and concatenated into a prompt $p$, which we used for inference over all examples  $x_{test}$ in the test set.
As an alternative to creating prompts, there is now a growing interest in customizing the in-context examples each example $x_{test}$. 
Popular techniques typically train a retriever, which is used to fetch the closest examples~\citep{liu_what_2021,rubin_learning_2021,poesia2021synchromesh}.
We also experimented with such \emph{dynamic} creation of the prompt, that depends on the particular test example.
Specifically, following ~\citet{poesia2021synchromesh}, we performed knowledge similarity tuning~(\kst): we trained a retriever model to retrieve the $k$ closest examples for a given input.

\begin{table}[h!]
\centering
\small
\begin{tabular}{llll}
\toprule
      Setup           & p & r & \fscore \\ \midrule
\ours & 57.34 & 55.44 & 56.52   \\
\ours + \kst & \textbf{67.11} & \textbf{64.57} & \textbf{65.71}   \\ \bottomrule
\end{tabular}
\caption{Our retrieval mechanism is highly effective for edge prediction: the closest examples are from similar domains and the model is able to leverage the information for better performance. }
\label{tab:proscriptedgepredkst}
\end{table}

The results indicate that the efficacy of dynamic prompts depends on both the training data and task. 
In the edge-prediction sub-task of \proscr , edges between events in similar scripts are helpful, and Table~\ref{tab:proscriptedgepredkst} shows that the model was able to effectively leverage this information.
In the script generation sub-task of \proscr,  Table~\ref{tab:proscriptkstappendix} shows that \kst provides  gains as well (Appendix~\ref{sec:dynamicpromptcreationappendix}). 

In \explg, we observed that the training data had multiple examples which were nearly identical, and thus dynamically created prompts often included such duplicate examples, effectively reducing diversity and prompt size~(\Cref{tab:explagraphkstappendix}).

\paragraph{Python Formatting}
We performed an extensive study of the effect of the Python format on the downstream task performance in Appendix~\ref{sec:pythonformatting}.
We find that: \begin{inparaenum}[(i)]
\item there are no clear task-agnostic Python class designs that work uniformly well; and that \item larger models are less sensitive to prompt (Python class) design. In general, our approach benefits the most from code formats that as similar as possible to the conventions of typical code.
\end{inparaenum}





\paragraph{Human evaluation} \emnlpcr{We conduct human evaluation of the graphs generated by \ours and \davinci to supplement automated metrics.
The results (\Cref{sec:humanevaluation}) indicate that human evaluation is closely correlated with the automated metrics: for \explg, graphs generated by \ours are found to be more relevant and correct.
For \proscr generation, both \davinci and \ours have complementary strengths, but \ours is generally better in terms of relevance.}
\section{Related work}





\paragraph{Structured commonsense reasoning using LLMs}
Existing methods for structured commonsense generation typically flatten the output graphs as strings~\citep{madaan2020neural,madaan2021could,sakaguchi_proscript_2021}. Consequently, these methods struggle with generation of well-formed outputs~\citep{sakaguchi_proscript_2021,madaan2021think}. In contrast, we address the problem of structured generation by  \begin{inparaenum} [(1)] \item translating the task into Python code, and \item generating code using large-code generation models. \end{inparaenum}

\paragraph{Code representation for procedural knowledge reasoning}
Programs inherently encode rich structures, and they can efficiently represent task procedures. Existing works leverage the control-flows, nested functions and API calls of a programming language such as Python to control the situated agents in the embodied environment~\citep{sun2019program,zhou22suki,singh2022progprompt}. In this work, we go beyond these procedural tasks and show the effectiveness of using \clm on broader structured commonsense tasks. 

\paragraph{Adapting \clm for reasoning}
As code-generation models~(\clm) are getting increasingly popular, there is a growing interest in adapting them for a wide range reasoning tasks.
\citet{wu2022autoformalization} use \codex and PaLM~\citep{chowdhery2022palm} for converting mathematical statements written in natural language into a formal structure that can be used for theorem provers, with moderate success.
The task is challenging, as it involves understanding the concepts used in the theorem~(\eg set of real numbers) and the complex relationship between them.
Our work is similar in spirit to \citet{wu2022autoformalization}, and seeks to leverage the dual abilities of \clm for text and symbolic reasoning.
However, differently from their work, we close the gap between the pre-training data and our tasks by translating our output into Python code.
As our experiments show, this step is crucial in outperforming text-only and fine-tuned models.
To the best of our knowledge, our work is the first to transform a natural-language reasoning problem into code to successfully leverage code generation methods.

\paragraph{Symbolic reasoning using LLMs}
\emnlpcr{The use of programming languages like LISP~\citep{tanimoto1987elements} and Prolog~\citep{colmerauer1996birth} to process natural language has a long history in AI.}
\emnlpcr{However, the recent progress in large language models has obviated the need for specialized methods for symbolic processing.}
\citet{cobbe2021training} and \citet{chowdhery2022palm} address middle-school level algebra problem solving using large-language models in a few-shot setup.
These problems require a model to understand the order in which a set of operations should be performed over symbols~(typically small integers).
In contrast, structured commonsense reasoning requires \emph{broader} information than supplied in the prompt, while utilizing the models' structural generation capabilities for generating output effectively.
Thus, the tasks in our work push a model to use \emph{both} its reasoning and symbolic manipulation capabilities.

\section{Conclusion}

We present the first work to employ large language models of code for structured commonsense generation.
By converting the output commonsense structures to Python code, \ours provides a simple and effective method for leveraging the code-generation abilities of \clm for structured generation.
These results open a promising direction for structural commonsense reasoning.
We believe that the principles and the methods presented
in this paper are  applicable to additional NLP tasks that require ``language understanding'' \emph{and} structured prediction.

\section*{Acknowledgments}
We thank Kaixin Ma, Keisuke Sakaguchi and Niket Tandon for thoughtful discussion and helping with \proscr datasets and the anonymous reviewers for valuable feedback. 
This material is partly based on research sponsored in part by the Air Force Research Laboratory under agreement number FA8750-19-2-0200. 
The U.S. Government is authorized to reproduce and distribute reprints for Governmental purposes notwithstanding any copyright notation thereon. 
The views and conclusions contained herein are those of the authors and should not be interpreted as necessarily representing the official policies or endorsements, either expressed or implied, of the Air Force Research Laboratory or the U.S. Government.
This project was also partially supported by a gift from AWS AI.
\section*{Limitations}
Some experiments in this work are performed with language models that are not open-sourced, namely \davinci, \curie, and \codex. 
Existing documentation~\citep{brown2020language,chen2021evaluating} does not fully describe the details of these models, such as the pretraining corpus, model size, and model biases.
Therefore, we can only provide educational guesses on these details~(analysis in Appendix~\ref{sec:modeldescription}). 
In addition, even though \codex is free to use for research as of June 2022, we are unsure whether the research community will continue to have free access in the future.
Nonetheless, we release our code and model outputs to ensure the reproducibility of our work. Furthermore, in cases where the models we experiment with reveal any issue, the publicly available code will allow future investigations.

\emnlpcr{Another limitation of our work is that we exclusively experiment with datasets in English. Exploring the efficacy of structured generation methods in cross-lingual settings is an interesting and important future work.}

\bibliography{custom}
\bibliographystyle{acl_natbib}
\newpage
\clearpage
\appendix
\section{Few-shot models size estimates}
\label{sec:modeldescription}

As OpenAI has not released any details of the size of their few-shot models, we estimate the relative strengths and weaknesses on code and text generation by calculating the average loss per token. 
To calculate the avg. loss of each of these models on code, we use the implementation provided by~\citet{xu2022systematic}.\footnote{\url{https://github.com/VHellendoorn/Code-LMs\#evaluation}}
The perplexity on text corpus was evaluated on 30 random wikipedia pages from Wikiplots\footnote{\url{https://github.com/markriedl/WikiPlots}} following a similar procedure
The structure and text generation capabilities of the models are apparent from the results in Table~\ref{tab:ppleval}; \davinci outperforms \codex on text generation but is worse on code-generation and vice-versa.
\curie underperforms both \davinci and \codex significantly.
Importantly, these results show that \codex and \davinci are of comparable capacities, making their comparison fair.

\begin{table}[!ht]
\centering
\begin{tabular}{@{}lrr@{}}
\toprule
Model &  \textsc{code}  & \textsc{text}  \\ \midrule
 \codex      &    \textbf{0.46}    &  2.71  \\
\davinci      &   0.63 &    \textbf{2.25}  \\
 \curie     &  1.17    &    3.32          \\
 \bottomrule
\end{tabular}
\caption{Average loss per token of the three few-shot models used in this work. \textsc{text} refers to the average loss over 30 Wikipedia pages, and \textsc{code} is the loss over Python scripts in the evaluation split of Polycoder.}
\label{tab:ppleval}
\end{table}

\section{Dynamic prompt Creation}
\label{sec:dynamicpromptcreationappendix}

\newcommand{\txt}{\mathcal{T}}
As an alternative to creating prompts, there is now a growing interest in customizing the in-context examples each example $\txt_{test}$. 
Popular techniques typically train a retriever, which is used to fetch the examples in the training set that are closest to $\txt_{test}$~\citep{liu_what_2021,rubin_learning_2021,poesia2021synchromesh}.

Specifically~\citet{poesia2021synchromesh} train a retriever with a \textit{target-similarity tuning}~(TST) objective over a corpus of $\mathcal{D}$ of $(x, y)$ examples.
TST learns an embedding function $f$ such that for a pair of examples $(x_i, y_i)$ and $(x_j, y_j)$, if $y_i \sim y_j \implies f(x_i) \sim f(x_j)$.
For a new $x$, $f(x)$ is used to retrieve the closest examples from $\mathcal{D}$.

We follow~\citet{poesia2021synchromesh}, and train a knowledge-similarity tuner~(\kst).
We use mpnet-base\footnote{\url{https://huggingface.co/microsoft/mpnet-base}} with SentenceTransformers~\citep{reimers-2019-sentence-bert} to fine-tune a retrieval function $f$ by minimizing the following loss:
\begin{align}
    L_{\theta} &= (\cos(f_{\theta}(\txt_i), f_{\theta}(\txt_j)) - \texttt{sim}(\struct_i, \struct_j))^2
\end{align}
where $f_{\theta}$ is parameterized using a transformer.

Results on using \kst with \proscr~(\Cref{tab:proscriptkstappendix}) and \explg~(\Cref{tab:explagraphkstappendix}). 
While \kst is highly effective for edge-prediction~\ref{tab:proscriptedgepredkst}, the results are mixed for \explg and \proscr.
For \proscr, \kst yields marginal gains. 
However, for \explg, a number of training examples have overlapping theme~(\Cref{tab:explagraphpromptexamples}), and thus creating a prompt dynamically reduces the effective information in the prompt.

\begin{table*}
\small
\centering
\begin{tabular}{@{}lrrrrrrrrrr@{}}
\toprule
& \iso & \ged & \avgdegree & \avgnumnodes & \avgnumedges  & \bleu & \rougel & \bleurt \\ \midrule
$\refscr$ & 1.0  & 0.0  & 1.84       & 7.41         & 6.8  & - & -  & -   & -        \\
\ours + 002~(15)    &   0.53 &	2.1	& 1.79 &	\textbf{7.44} &	\textbf{6.7} & 25.24 & 38.28  & -0.26       \\ 
\ours + 002~(15) + \kst   &   0.52 & 1.99	& \textbf{1.8} &	7.45 &	\textbf{6.7} & \textbf{25.4} & \textbf{38.4}  & \textbf{-0.25}       \\ 
\bottomrule
\end{tabular}
\caption{\kst on \proscr generation: Dynamically creating a prompt leads to marginal improvements.}
\label{tab:proscriptkstappendix}
\end{table*}

\begin{table*}
\centering
\begin{tabular}{llllll}
\hline
            & \stca~($\uparrow$) & \seca~($\uparrow$) & \gbs~($\uparrow$)  & \ged~($\downarrow$)  & \ea~($\uparrow$)   \\ \midrule
\ours + 002 & \textbf{45.2}& 	\textbf{23.74}& 	\textbf{34.68}& 	\textbf{68.76}& 	\textbf{23.58} \\ 
\ours + 002 + \kst & 37.47 &	18.46 &	29.41 &	73.76 &	19.15 \\ \bottomrule
\end{tabular}
\caption{\kst on \explg: We find that \explg contains multiple examples that are similar to each other in the training set. Thus, dynamically creating a prompt by selecting examples that are closest to the input actually hurts performance.}
\label{tab:explagraphkstappendix}
\end{table*}

\begin{table*}
\centering
\small
\begin{tabular}{@{}p{12cm}@{}}
\toprule
\textit{belief}: all religions need to be respected, and able to practice. \textit{argument}: religion is behind many wars.\\
\textit{belief}: every religion needs to be respected and allowed to be practiced. \textit{argument}: religion is behind most wars.\\
\textit{belief}: school prayer should not be allowed. \textit{argument}: many people would prefer to keep religion out of their lives.\\
\textit{belief}: people should follow whichever religion they choose. \textit{argument}: this country has freedom of religion.\\
\textit{belief}: people are free to practice the religion they choose\textit{argument}: society's right to be free to practice religion should not be limited.\\
\textit{belief}: the church of scientology should be allowed, because everyone has a right to follow their religion. \textit{argument}: the church of scientology doesn't have a religious doctrine.\\
\textit{belief}: we should avoid discussing religion in schools.\textit{argument}: some schools are religious in nature, and have regular discussions on the topic.\\
\textit{belief}: freedom of religion is paramount. \textit{argument}: not all religions are worth it.\\
\textit{belief}: people don't follow the same religion. \textit{argument}: the world has many different religions.\\
\textit{belief}: people should follow whatever religion they desire. \textit{argument}: people have the right to adhere to the religion of their choice.\\
\textit{belief}: people should follow whichever religion they choose. \textit{argument}: some religions are better than others.\\
\textit{belief}: people should be able to practice whatever religion they choose. \textit{argument}: some religions are not okay to pursue.\\
\textit{belief}: students have a right to express themselves any way possible, including faith. \textit{argument}: religion is a personal choice.\\
\textit{belief}: people should be able to do missionary work if they desire. \textit{argument}: people should have right to missionary work.\\
\textit{belief}: students are free to express faith. \textit{argument}: one should go to church to express their religious beliefs.\\
\bottomrule
\end{tabular}
\caption{The closest examples in the training set corresponding to the test input: \textit{belief}: religion causes many fights. and \textit{argument}: \textit{There would be less fights without religious conflicts.}. As the table shows, the examples are overlapping which reduces the diversity in the prompt, effectively reducing the number of examples in a prompt creating using nearest neighbors~(\Secref{sec:analysis}.}
\label{tab:explagraphpromptexamples}
\end{table*}

\section{Human Evaluation}
\label{sec:humanevaluation}

Out of the four tasks used in this work, \proscr edge prediction and \propar have only one possible correct value.
Thus, following prior work, we report the automated, standard metrics for these tasks.
For \explg, we use model-based metrics proposed by~\citet{saha_explagraphs_2021}, which were found to have a high correlation with human judgments. For \proscr graph generation, we conducted an exhaustive automated evaluation that separately scores the correctness of the nodes and the correctness of the edges.

However, automated metrics are limited in their ability to evaluate model-generated output. Thus, to further investigate the quality of outputs, we conduct a human evaluation to compare the outputs generated by \ours and \davinci. We sampled 20 examples, and three of the authors performed the evaluation.
Annotators were shown two graphs (generated by \ours and \davinci) and were asked to select one they thought was better regarding relevance and correctness.
The selection for each criterion was made independently: the same graph could The annotations were done separately: the same graph could have more relevant nodes (higher relevance) but may not be correct.
The identity of the model that generated each graph (\ours or \davinci) was shuffled and unknown to the evaluators.

\begin{table*}[]
\centering
\begin{tabular}{@{}lllll@{}}
\toprule
                             & Dataset                     & \ours & \davinci & No preference \\ \midrule
\multirow{2}{*}{Relevance}   & \explg                      & 28.3\%  & 16.7\%     & 46.7\%         \\
                             & \proscr (script generation) & 26.7\%  & 18.3\%    & 55\%           \\ \midrule
\multirow{2}{*}{Correctness} & \explg                      & 38.3\%  & 18.3\%     & 31.7\%         \\
                             & \proscr (script generation) & 26.7\%  & 23.3\%     & 50\%          \\ \bottomrule
\end{tabular}
\caption{Human evaluation of graphs generated by \ours and \davinci. The evaluators were shown graphs generated by \ours and \davinci, and were asked to select one that is more relevant to the input and correct. In case of no preference, the evaluators could pick the No preference. The table shows the \% of times graphs from each model were preferred.}
\label{tab:human-eval}
\end{table*}

The results in Table~\ref{tab:human-eval} indicate that human evaluation is closely correlated with the automated metrics: for \explg, annotators found the graphs generated by \ours to be more relevant and correct. We find that \davinci often fails to recover semantic relations between nodes in the argument graphs. 
For example, consider a belief (B) \textit{urbanization harms natural habitats for the animals in the world}. We want to generate a graph that can \textbf{counter} this belief with the argument (A) \textit{urbanization causes increase in jobs.}

For the same prompt, \ours generated \textit{(urbanization; causes; increase in jobs); (increase in jobs; has context; good); (good; not capable of; harms)} whereas \davinci generated \textit{(jobs; not harms; natural habitats) $\rightarrow$ (natural habitats; not part of; animals).} 
Note that \davinci successfully recovered relevant events (``natural habitat'' ``animals'') but arranged them in incorrect relations.
For \proscr, the human evaluation shows that \ours and \davinci have complementary strengths, while \ours generally produces more relevant and correct outputs.

\section{Dataset statistics}
\label{sec:appendixdataset}

Dataset statistics are shown in Table~\ref{tab:data_stats}. The test split for \explg is not available, so we evaluate on the validation split.
For \proscr, we obtained the test splits from the authors.

\begin{table}[H]
\centering
\begin{tabular}{@{}lccc@{}}
\toprule
\textbf{Corpus} & \textbf{\#Train} & \textbf{\#Val} & \textbf{\#Test}\\ \midrule
\proscr   & 3252 & 1085 &  2077          \\
\propar & 387 & 43 & 54   \\
\explg      & 2368 & 398 & -    \\ \bottomrule
\end{tabular}
\caption{Corpus Statistics for the tasks used in this work.} 
\label{tab:data_stats}
\end{table}

\section{Sample outputs}
\begin{figure*}[!ht]
\centering
{\includegraphics[width=0.48\textwidth,height=0.25\textheight]{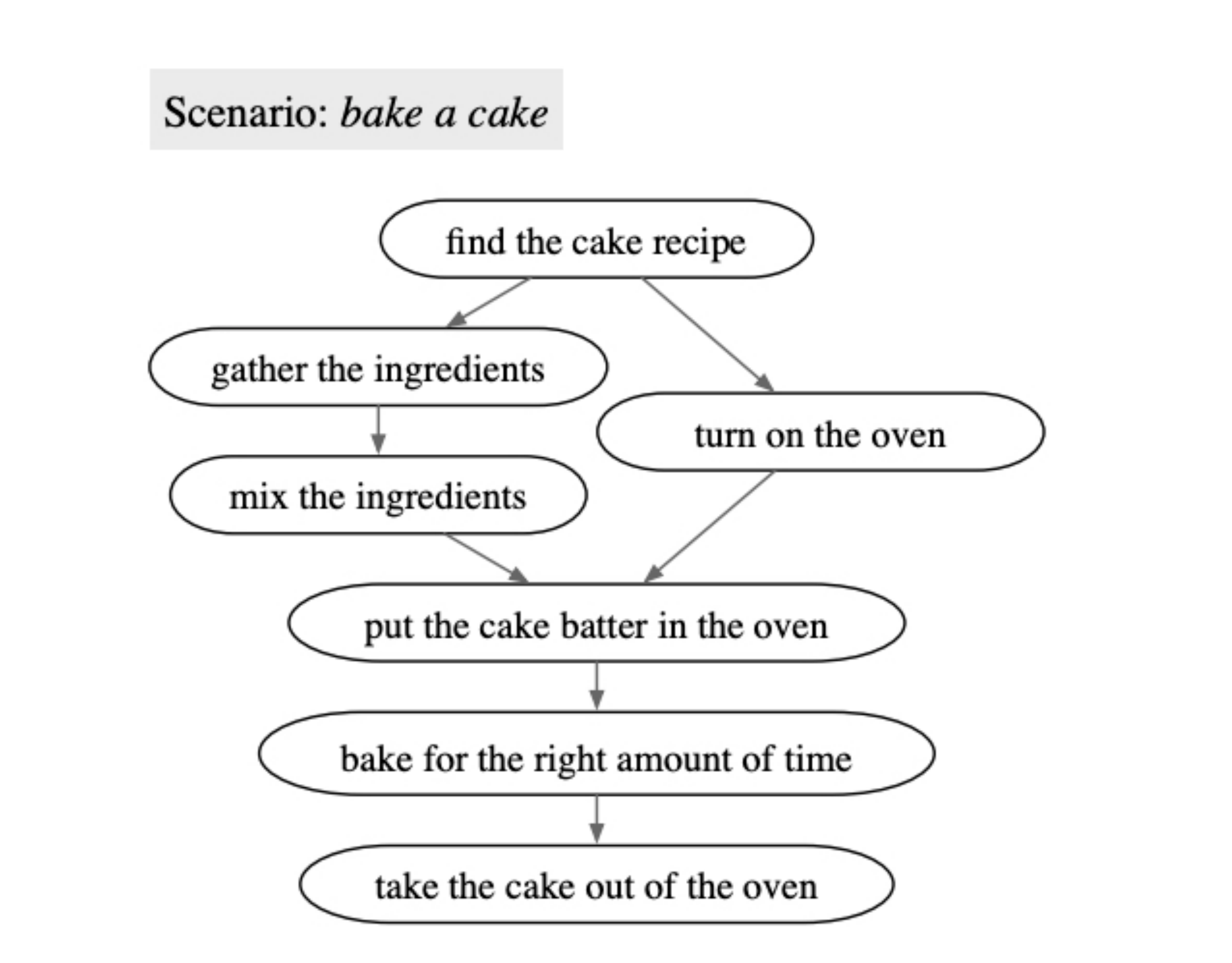}}
{\includegraphics[width=0.48\textwidth,height=0.25\textheight]{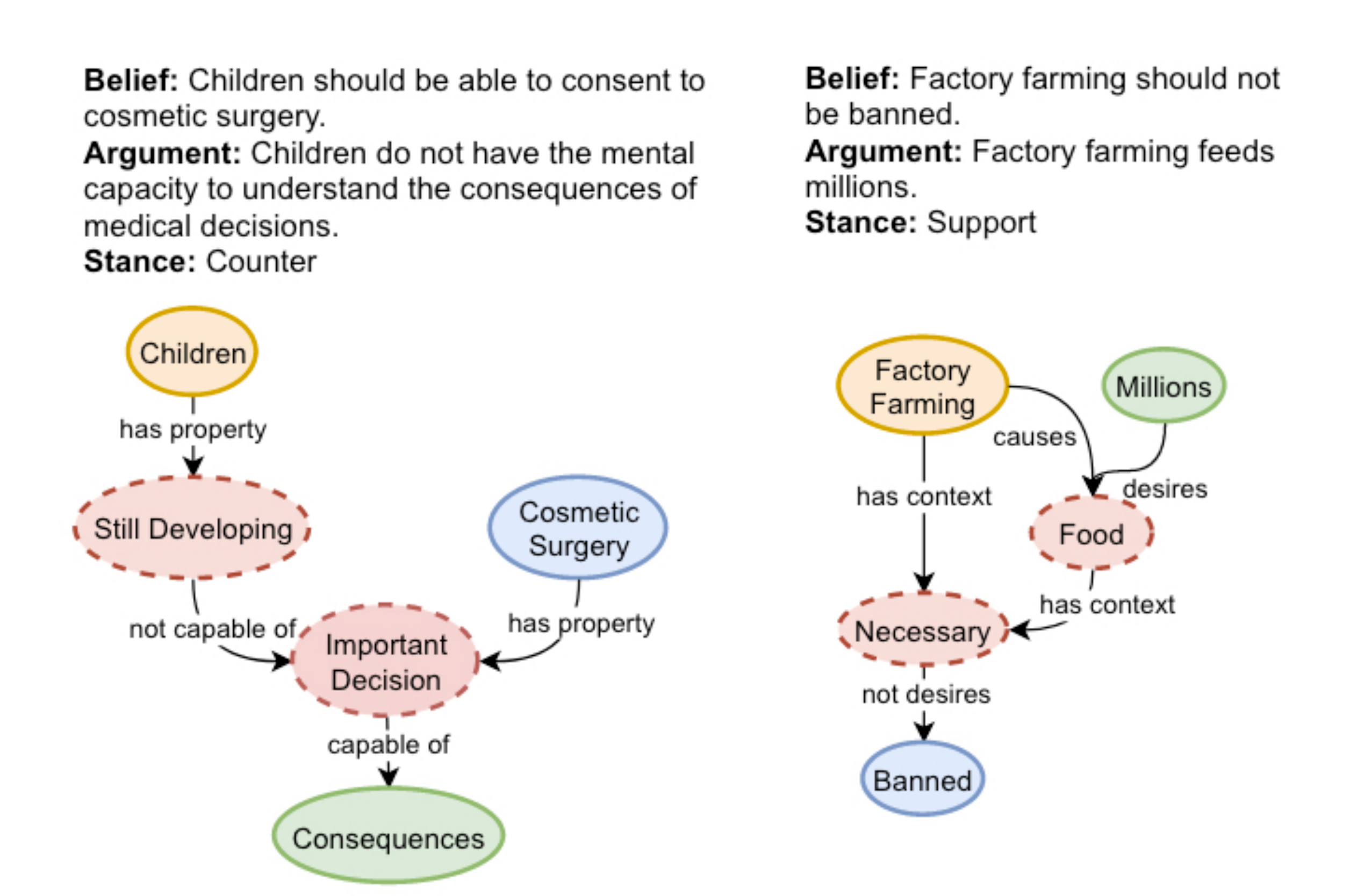}}
{\includegraphics[width=0.48\textwidth,height=0.15\textheight]{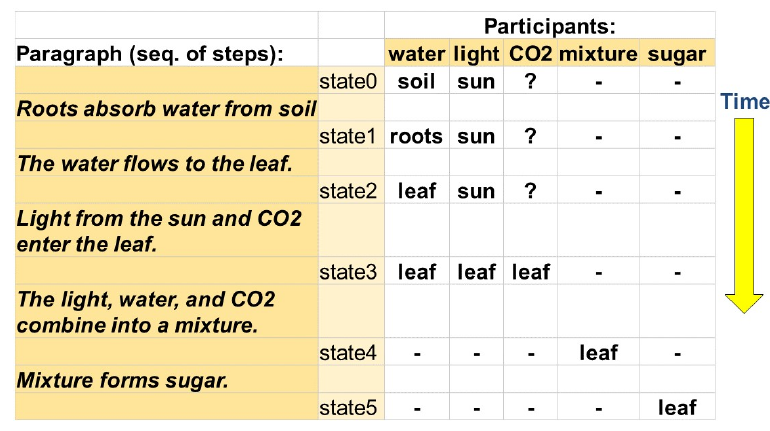}}
\caption{Example graphs for each of the tasks used for \ours: \proscr (top-left), \explg (top-right), and \propar(bottom).}
\label{fig:examples}
\end{figure*}

Sample outputs from \ours for all the tasks are located at~\url{https://github.com/madaan/CoCoGen/tree/main/outputs}.
Representative examples from each task are presented in Figure~\ref{fig:examples}.
Surprisingly, \ours~(\codex with a Python prompt) generates syntactically valid Python graphs that are similar to the task graphs/tables in nearly 100\% of the cases.

\begin{figure*}[!h]
  \begin{minipage}{0.95\textwidth}
  \centering
    \begin{tikzpicture}[node distance=1.5cm]
\newcommand{\nodespacing}{3pt}
\newcommand{\roundness}{2mm}
\newcommand{\intromarkersize}{1.5mm}
    
    \node [draw, rounded corners=\roundness,inner sep=\nodespacing] (node0) {Take the pies out to cool};
    \node [draw, right of=node0, xshift=2.5cm,rounded corners=\roundness,inner sep=\nodespacing]  (node1) {Open cabinet drawer};
    \node [draw, below of=node0, xshift=2cm,rounded corners=\roundness,inner sep=\nodespacing] (node2) {Take out several plates};
    \node [draw, below of=node2, xshift=-2cm,rounded corners=\roundness*2,inner sep=\nodespacing, text width=3cm,align=center] (node3) {Begin putting pies on plate};
    \node [draw, right of=node3, xshift=2.5cm,rounded corners=\roundness*2,inner sep=\nodespacing, text width=3cm,align=center] (node4) {Fill pies onto plates evenly};
    \node [draw, below of=node3, xshift=2cm,rounded corners=\roundness,inner sep=\nodespacing] (node5) {Serve the potpies on a plate};
    \draw [-{Latex[length=\intromarkersize, line width=1pt]}] (node0) -- (node2);
    \draw [-{Latex[length=\intromarkersize, line width=1pt]}] (node1) -- (node2);
    \draw [-{Latex[length=\intromarkersize, line width=1pt]}] (node2) -- (node3);
    \draw [-{Latex[length=\intromarkersize, line width=1pt]}] (node2) -- (node4);
    \draw [-{Latex[length=\intromarkersize, line width=1pt]}] (node3) -- (node5);
    \draw [-{Latex[length=\intromarkersize, line width=1pt]}] (node4) -- (node5);
\end{tikzpicture}
\end{minipage}
\vspace{1em}
  \begin{minipage}{0.95\textwidth}
   \centering
 \begin{minted}[fontsize=\footnotesize,framesep=2pt,frame=single,autogobble]{python}
 class Tree:
 
   goal = "serve the potpies on a plate"
 
   def __init__(self):
     # nodes
     begin = Node()
     take_pies_out_to_cool = Node()
     take_out_several_plates = Node()
     open_cabinet_drawer = Node()
     fill_pies_onto_plates_evenly = Node()
     begin_putting_pies_on_plates = Node()
     serve_potpies_on_plate = Node()
 
     # edges
     begin.children = [take_pies_out_to_cool, open_cabinet_drawer]
     take_pies_out_to_cool.children = [take_out_several_plates]
     open_cabinet_drawer.children = [take_out_several_plates]
     take_out_several_plates.children = [begin_putting_pies_on_plates,
         fill_pies_onto_plates_evenly]
     begin_putting_pies_on_plates.children = [serve_potpies_on_plate]
     fill_pies_onto_plates_evenly.children = [serve_potpies_on_plate]
     serve_potpies_on_plate.children = [end]
   \end{minted}
 
  \end{minipage}
  \captionof{figure}{A \proscr plan (top) and the corresponding Python code (bottom).}
   \label{fig:proscript_example_complete}

 \end{figure*}

\section{Prompts}
\label{sec:appendixprompts}

The prompts for each tasks are present at this anonymous URL:

\begin{enumerate}
    \item \proscr script-generation: \url{https://github.com/madaan/CoCoGen/tree/main/data/proscript_script_generation/prompt.txt}
    \item \proscr edge-prediction: \url{https://github.com/madaan/CoCoGen/tree/main/data/proscript_edge_prediction/prompt.txt}
\item \propar: \url{https://github.com/madaan/CoCoGen/tree/main/data/explagraphs/prompt.txt}
    
    \item \explg: \url{https://github.com/madaan/CoCoGen/tree/main/data/explagraphs/prompt.txt}
    
\end{enumerate}

These prompts are also present in the attached supplementary material, and can be found in the \code{data} folder under respective task sub-directories.

 \section{Designing Python class for a structured task}
 \label{sec:pythonformatting}
 
Figure~\ref{fig:explagraph_templates} shows three different designs for Explagraphs.
For \proscr, the various formats include representing proscript as a Networkx\footnote{\url{https://networkx.org/}} class~(\ref{fig:proscript_networkx}), DOT-like class~\ref{fig:proscript_dot}, and as a Tree~(\ref{fig:proscript_tree}).

 \begin{table*}[]
\centering
\begin{tabular}{lllllll}
\toprule
Model      & Format   & \stca~($\uparrow$) & \seca~($\uparrow$) & \gbs~($\uparrow$)  & \ged~($\downarrow$)  & \ea~($\uparrow$)     \\ \midrule
\codex-002 & Literal  & \textbf{45.2}  & \textbf{23.74} & \textbf{34.68} & \textbf{68.76} & \textbf{23.58} \\
\codex-002 & Tree     & 39.24 & 15.95 & 30.49 & 73.85 & 18.24 \\
\codex-002 & Relation & 42.82 & 23.68 & 33.38 & 70.23 & 21.16 \\ \bottomrule
\end{tabular}
\caption{Performance of \codex on the three different formats present in Figure~\ref{fig:explagraph_templates} for \explg.}
\label{tab:explagraphformatsensitivity}
\end{table*}

 \begin{table}[]
\centering
\begin{tabular}{lllllll}
\toprule

Model      & Format   & \fscore     \\ \midrule
\codex-001 & Literal  & 15.9 \\
\codex-001 & Tree     &  29.7    \\
\codex-002 & Literal~(\Cref{fig:proscript_dot})  & 52.0  \\
\codex-002 & Tree~(\Cref{fig:proscript_tree})     & 56.5 \\ \bottomrule
\end{tabular}
\caption{Performance of \codex-001 and \codex-002 on the the different formats present in Figure~\ref{fig:proscript_tree} and \ref{fig:proscript_dot} for \proscr edge prediction. We find that the literal format that combines structure with literally Figure output performs the best for \codex-002.}
\label{tab:edgepredformatsensitivity}
\end{table}

\begin{table*}
\small
\centering
\begin{tabular}{llrrrrrrrrr}
\toprule
  Model & Format        & \iso & \ged& \avgdegree & \avgnumnodes & \avgnumedges  & \bleu & \rougel &  \bleurt \\ \midrule
$\refscr$ & - &1.0  & 0.0  & 1.84       & 7.41         & 6.8  & - & -  & -          \\
\codex-001 & Literal~(\Cref{fig:proscript_dot})  & \textbf{0.55}&	\textbf{1.8} &	1.74&	7.45&	6.5    &   22.9 &	36.2 &		-0.36      \\ 
\codex-001 & Tree~(\Cref{fig:proscript_tree})  &  0.35 &	3 &	\textbf{1.79} &	7.45 &	6.65 &		17.8&	30.7&	-0.45      \\ 
\codex-001 & NetworkX~(\Cref{fig:proscript_networkx})  & 0.51 &	1.81  &	1.69 &	7.49 &	6.32 & 23.7& 	35.9&  -0.37     \\ 
\codex-002 & Literal~(\Cref{fig:proscript_dot})  &  0.53 &	2.1		& \textbf{1.79} &	\textbf{7.44} &	\textbf{6.7} & \textbf{25.24} & \textbf{38.28}  &  \textbf{-0.26}       \\ 
\codex-002 & Tree~(\Cref{fig:proscript_tree})   &  0.35	 &2.46	 &1.61	 &7.46	 & 5.74		&18.96&	32.92&	-0.38        \\ 
\codex-002 & NetworkX~(\Cref{fig:proscript_networkx})  & 0.5& 	2.46& 	\textbf{1.79}& 	\textbf{7.38}& 	6.61 & 23.88 &	36.89 &	-0.33    \\ 
\bottomrule
\end{tabular}
\caption{\codex results on \proscr generation for various Python source formats.}
\label{tab:proscriptformatsensitivity}
\end{table*}

\begin{figure*}
   \begin{minipage}{0.95\textwidth}
   \centering
 \begin{minted}[fontsize=\footnotesize,framesep=2pt,frame=single,autogobble]{python}
 


class Relation:

    def __init__(self):
        belief = "Cannabis should be legal."
        argument = "It's not a bad thing to make marijuana more available."
        stance = "support"

        # create a DAG to support belief using argument
        begin = ["cannabis"]
        add_edge("cannabis", "synonym of", "marijuana")
        add_edge("legal", "causes", "more available")
        add_edge("marijuana", "capable of", "good thing")
        add_edge("good thing", "desires", "legal")

class Tree:
    def __init__(self):
        self.belief = "Cannabis should be legal."
        self.argument = "It's not a bad thing to make marijuana more available."
        self.stance = "support"

        # tree for support in support of belief
        root_nodes = cannabis
        cannabis = Node()
        cannabis.add_edge("synonym of", "marijuana")
        legal = Node()
        legal.add_edge("causes", "more available")
        marijuana = Node()
        marijuana.add_edge("capable of", "good thing")
        good_thing = Node()
        good_thing.add_edge("desires", "legal")


 class Literal:
    def __init__(self):
        self.belief = "Cannabis should be legal."
        self.argument = "It's not a bad thing to make marijuana more available."
        self.stance = "support"
        self.graph = """\
(cannabis; synonym of; marijuana)(legal; causes; more available)
(marijuana; capable of; good thing)
(good thing; desires; legal)"""




   \end{minted}

  \end{minipage}
  \captionof{figure}{Templates tried for explagraph.}
   \label{fig:explagraph_templates}

 \end{figure*}

\begin{figure*}
   \begin{minipage}{0.95\textwidth}
   \centering
 \begin{minted}[fontsize=\footnotesize,framesep=2pt,frame=single,autogobble]{python}
 
class Plan:

    goal = "create a video game"
    num_steps = 7

    def __init__(self):
        graph = nx.DiGraph()
        # add nodes
        step0 = "decided to create a video game"
        step1 = "Learn the basics of programming"
        step2 = "Learn to use a language that is used in games"
        step3 = "Learn to use an existing game engine"
        step4 = "Program the game"
        step5 = "Test the game"
        step6 = "create a video game"
        graph.add_nodes_from([step0, step1, step2, step3, step4, step5, step6])

        # add edges
        graph.add_edge(step0, step1)
        graph.add_edge(step1, step2)
        graph.add_edge(step1, step3)
        graph.add_edge(step2, step4)
        graph.add_edge(step3, step4)
        graph.add_edge(step4, step5)
        graph.add_edge(step5, step6)


   \end{minted}

  \end{minipage}
  \captionof{figure}{Proscript as a Networkx class.}
   \label{fig:proscript_networkx}

 \end{figure*}

\begin{figure*}
   \begin{minipage}{0.95\textwidth}
   \centering
 \begin{minted}[fontsize=\footnotesize,framesep=2pt,frame=single,autogobble]{python}
 class CreateAVideoGame:

    title = "create a video game"
    steps = 7

    def step0(self):
        return "decided to create a video game"
    def step1(self):
        return "Learn the basics of programming"
    def step2(self):
        return "Learn to use a language that is used in games"
    def step3(self):
        return "Learn to use an existing game engine"
    def step4(self):
        return "Program the game"
    def step5(self):
        return "Test the game"
    def step6(self):
        return "create a video game"
    def get_relations(self):
        return [
            "step0 -> step1",
            "step1 -> step2",
            "step1 -> step3",
            "step2 -> step4",
            "step3 -> step4",
            "step4 -> step5",
            "step5 -> step6",
        ]


   \end{minted}

  \end{minipage}
  \captionof{figure}{Representing \proscr graph literally.}
   \label{fig:proscript_dot}

 \end{figure*}

 \begin{figure*}
   \begin{minipage}{0.95\textwidth}
   \centering
 \begin{minted}[fontsize=\footnotesize,framesep=2pt,frame=single,autogobble]{python}

class Tree:

    goal = "serve the potpies on a plate"

    def __init__(self):
        # nodes
        begin = Node()
        take_pies_out_to_cool = Node()
        take_out_several_plates = Node()
        open_cabinet_drawer = Node()
        fill_pies_onto_plates_evenly = Node()
        begin_putting_pies_on_plates = Node()
        serve_potpies_on_plate = Node()

        # edges
        begin.children = [take_pies_out_to_cool, open_cabinet_drawer]
        take_pies_out_to_cool.children = [take_out_several_plates]
        open_cabinet_drawer.children = [take_out_several_plates]
        take_out_several_plates.children = [begin_putting_pies_on_plates,
                                            fill_pies_onto_plates_evenly]
        begin_putting_pies_on_plates.children = [serve_potpies_on_plate]
        fill_pies_onto_plates_evenly.children = [serve_potpies_on_plate]
        serve_potpies_on_plate.children = [end]
   \end{minted}

  \end{minipage}
  \captionof{figure}{Proscript with a tree-encoding.}
   \label{fig:proscript_tree}

 \end{figure*}

\section{Impact of Model size}
\label{sec:modelsize}
The \codex model released by OpenAI is available in two versions\footnote{as of June 2022}: 
\code{code-davinci-001} and \code{code-davinci-002}.
While the exact sizes of the models are unknown because of their proprietary nature, OpenAI API states that \code{code-davinci-002} is the \textit{Most capable Codex model}
Tables~\ref{tab:proscriptresultsacrossmodelsize} and \ref{tab:explagraphresultsacrossmodelsize} compares \ours+\code{code-davinci-001} with \ours+\code{code-davinci-002}.
Note that both \code{code-davinci-001} and \code{code-davinci-002} can fit ~4000 tokens, so the number of in-context examples was identical for the two settings.
The results show that for identical prompts, \ours+\code{code-davinci-002} vastly outperforms \ours+\code{code-davinci-001}, showing the importance of having a better underlying code generation model.

\begin{table*}
\small
\centering
\begin{tabular}{lrrrrrrrrrr}
\toprule
          & \iso & \ged  & \avgdegree & \avgnumnodes & \avgnumedges  & \bleu & \rougel &  \bleurt \\ \midrule
$\refscr$ & 1.0  & 0.0  & 0.0      & 1.84       & 7.41         & 6.8  & - & -   & -        \\
\ours + 001~(15)    &   0.55&	1.8 &	1.74&	7.45&	6.5    &   22.9 &	36.2 &	-0.36      \\ 
\ours + 002~(15)    &   0.53 &	2.1		& 1.79 &	\textbf{7.44} &	\textbf{6.7} & \textbf{25.24} & \textbf{38.28}    & \textbf{-0.26}       \\ 
\bottomrule
\end{tabular}
\caption{\codex-001 vs 002 on \proscr script generation}
\label{tab:proscriptresultsacrossmodelsize}
\end{table*}



\begin{table*}[H]
\centering
\small
\begin{tabular}{llllll}
\toprule
& \stca~($\uparrow$) & \seca~($\uparrow$) & \gbs~($\uparrow$)  & \ged~($\downarrow$)  & \ea~($\uparrow$)   \\ \midrule
\ours + 001 & 33.16 & 	7.14& 	25.91& 	77.45& 	15.9  \\
\ours + 002 & \textbf{45.2}& 	\textbf{23.74}& 	\textbf{34.68}& 	\textbf{68.76}& 	\textbf{23.58} \\
\bottomrule
\end{tabular}
\caption{\codex-001 vs 002 on \explg.}
\label{tab:explagraphresultsacrossmodelsize}
\end{table*}

\paragraph{Model size vs. sensitivity to the prompt}
In Table~\ref{tab:edgepredformatsensitivity} shows the performance of \codex-001 (smaller) and \codex-002 (larger, also see Appendix~\ref{sec:modeldescription}) on identical prompts. Our experiments show that as model size increases, the sensitivity of the model on the prompt reduces. This indicates that for very large models, prompt design might get progressively easier.

\section{Variation in prompts}
\label{sec:variationprompts}

We run each experiment with 3 different random seeds, where the random seeds decides the order of examples in the prompt.
We find minimal variance between runs using different fixed prompts between 3 runs.
Further, as shown in the Tables~\ref{tab:proscript_scriptgen_variation},\ref{tab:proscript_edgepred_variation}, \ref{tab:explagraph_variation}, and \ref{tab:propar_variation}, all improvements of \ours over \davinci are statistically significant (p-value $<$ 0.001).

\begin{table}[H]
\centering
\begin{tabular}{@{}llll@{}}
\toprule
         & \bleu    & \rougel   & \bleurt    \\ \midrule
\davinci & 23.1$\pm$2.7 & 36.5$\pm$2.7 & -0.27$\pm$0.06 \\
\ours    & 25.3$\pm$0.1 & 38.3$\pm$0.1 & -0.25$\pm$0.01 \\ \bottomrule
\end{tabular}
\caption{\proscr script generation: mean and standard deviation across three different random seeds.}
\label{tab:proscript_scriptgen_variation}
\end{table}

\begin{table}[H]
\centering
\begin{tabular}{@{}ll@{}}
\toprule
         & F1                \\ \midrule
\davinci & 48.9 $\pm$ 2.8        \\
\ours    & \textbf{56.2$\pm$2.1} \\ \bottomrule
\end{tabular}
\caption{\proscr edge prediction: mean and standard deviation across three different random seeds.}
\label{tab:proscript_edgepred_variation}
\end{table}

\begin{table*}
\centering
\small
\begin{tabular}{llllll}
\toprule
& \stca~($\uparrow$) & \seca~($\uparrow$) & \gbs~($\uparrow$)  & \ged~($\downarrow$)  & \ea~($\uparrow$)   \\ \midrule
\davinci & 25.4 $\pm$ 2.7    &        13.7 $\pm$ 2.8   &         20 $\pm$ 2.3    &           82.5 $\pm$ 1.9       &     13.6 $\pm$ 1.8   \\
\ours  & \textbf{44.0} $\pm$ 1.2  & \textbf{25.1} $\pm$ 2.5 &  \textbf{34.1} $\pm$ 0.7 &  \textbf{69.5} $\pm$ 0.7 &  \textbf{22.0} $\pm$ 1.3  \\
\bottomrule
\end{tabular}
\caption{\explg: mean and standard deviation across three different random seeds.}
\label{tab:explagraph_variation}
\end{table*}

\begin{table}[H]
\centering
\begin{tabular}{@{}ll@{}}
\toprule
         & F1                \\ \midrule
\davinci & 56.9 $\pm$ 2.4        \\
\ours    & \textbf{62.8} $\pm$ 2.4  \\ \bottomrule
\end{tabular}
\caption{\propar: mean and standard deviation across three different random seeds.}
\label{tab:propar_variation}
\end{table}

\end{document}